\documentclass[a4paper, 10pt]{article}

\usepackage{packageList}
\usepackage{macros}

\graphicspath{{./Figures/}}
\pgfplotsset{compat=1.11}
\newlength\fwidth

\usepackage[linesnumbered]{algorithm2e}

\title{Finetuning greedy kernel models by exchange algorithms}
\author[1]{Tizian Wenzel \thanks{tizian.wenzel@uni-hamburg.de}}
\author[1]{Armin Iske \thanks{armin.iske@uni-hamburg.de}}
\affil[1]{Department of Mathematics, Universität Hamburg, \newline Bundesstr. 55, 20146, Hamburg, Germany}

\usepackage{cleveref}

\newif\iflong			
\longfalse

\begin{document}

\maketitle %
  
\begin{abstract}
Kernel based approximation offers versatile tools for high-dimensional approximation, which can especially be leveraged for surrogate modeling.
For this purpose, both ``knot insertion'' and ``knot removal'' approaches aim at choosing a suitable subset of the data,
in order to obtain a sparse but nevertheless accurate kernel model.

In the present work, focussing on kernel based interpolation,
we aim at combining these two approaches to further improve the accuracy of kernel models,
without increasing the computational complexity of the final kernel model. 
For this, we introduce a class of kernel exchange algorithms (KEA).
The resulting KEA algorithm can be used for finetuning greedy kernel surrogate models,
allowing for an reduction of the error up to 86.4\% (17.2\% on average) in our experiments.
\end{abstract}

\section{Introduction} \label{sec:introduction}

Kernel methods comprise versatile tools for multivariate numerical approximation, 
statistical learning and machine learning \cite{hofmann2008kernel, steinwart2008support, wendland2005scattered}.
They are popular due to their easy implementation, good performance and well-established mathematical theory based on reproducing kernel Hilbert spaces.
In applications, sparse kernel models are frequently used for surrogate modeling purposes \cite{doeppel2024goal, dutta2020greedy, haasdonk2023certified},
where efficient models are required that are able to accurately describe an expensive full model.
This can be achieved by using only a small subset of a possibly big initial training set for computing the final model.
For this, the computation of an optimal subset is frequently intractable due to the high combinatorial complexity \cite{hochbaum1997approximation}.
Thus, a more practical way is provided by using greedy algorithms, which are only \textit{locally optimal}, however very efficient to run.
For the task at hand, these greedy algorithms either start with a small or even empty set and then add points, or they start with the full set and then remove points.
In both approaches, a desired expansion size or a desired accuracy threshold may serve as stopping criteria.

In first approach, i.e.\ \textit{inserting} points, most work focussed on greedy algorithms that iteratively add single points according to some selection criterion.
For this, various criteria with different purposes and advantages have been introduced, e.g.\ 
the $f$-greedy algorihtm \cite{schaback2000adaptive}, 
the $P$-greedy algorithm \cite{marchi2005optimal}, or 
the $f / P$-greedy algorithm \cite{mueller2009komplexitaet}.
These algorithms were jointly analyzed in a framework of so-called $\beta$-greedy algorithms in \cite{wenzel2023analysis}.
Despite a greedy algorithm is a concatenation of local optimal steps,
the overall procedure usually does not give a globally optimal solution.
Nevertheless, some of these greedy insertion algorithms have been proven to be \textit{asymptotically} optimal:
For the $P$-greedy algorithm, the optimality of the resulting convergence rate in several cases was proven in \cite{santin2017convergence, wenzel2021novel},
while the optimality of the convergence rate of the $f$-greedy algorithm is discussed in \cite{santin2023optimality}.

The second approach, i.e.\ \textit{removing} points, was proposed in \cite{floater1996multistep, floater1998thinning} under the notion of \textit{thinning algorithms}, 
for the purpose of generating quasi-uniformly distributed subsets of scattered points.
Adaptive thinning algorithms were later suggested and analyzed in \cite{demaret2006image, demaret2006adaptive}.
Further work in this direction was done e.g.\ in \cite{marchetti2022efficient} under the notion \textit{knot removal} schemes,
where an \textit{efficient reduced basis algorithm} (ERBA) was introduced for removing points.
Also for these removal algorithms, optimality results are available \cite{demaret2015optimal}.

As these greedy insertion algorithms are (partly) known to be asymptotically optimal, 
the rate of convergence cannot be improved anymore.
Nevertheless it may be possible to further minimize the absolute error for a given expansion size, 
without achieving a faster convergence rate, i.e.\ by minimizing the prefactors in front of the asymptotic decay.
To the best of the authors knowledge, this subject was never considered in the literature so far,
probably because it is not possible to modify the greedy selection criteria such that a smaller error is obtained for \textit{any} expansion size.
In this sense, greedy algorithms can be seen as limited when aiming for a small error given an restricted expansion size (budget).

While one might be tempted to think about a global optimization of the centers and a decoupling of centers and function values (as in unsymmetric collocation \cite{Kansa1990a}),
this would likely require costly gradient descent techniques while also loosing the theoretical access based on the well-known kernel representer theorem \cite{fasshauer2015kernel, wendland2005scattered}.

Therefore we introduce \textit{exchange algorithms}, which solely make use of the available training data and thus stick to the framework and mathematical theory provided by the representer theorem.
By using an initial set of greedily selected centers -- obtained either via insertion or removal strategies --
and a subsequent exchange steps of these centers, we are able to finetune greedy kernel models.
While this introduces a small computational overhead, the improved accuracy of the kernel model frequently pays off in subsequent evaluations,
as the number of centers and thus the cost of evaluation of the kernel model stays fixed.
All in all, the combination of greedy algorithms and kernel exchange algorithms is a further contribution step towards the optimal selection of centers for kernel approximation.

The paper is structured as follows:
In \Cref{sec:review_greedy}, greedy kernel algorithms are reviewed,
which serve as building blocks for the kernel exchange algorithm (KEA),
which is introduced and discussed in \Cref{sec:kea}.
\Cref{sec:numerical_experiments} showcases the use of the KEA algorithms on a variety of use cases, 
achieving improvements of up to 86.4\%.
Finally \Cref{sec:conclusion_outlook} concludes the paper.

\section{Background on greedy kernel models}
\label{sec:review_greedy}

The following section reviews the most important terminology from kernel interpolation,
which is required for the introduction and discussion of the kernel exchange algorithms (KEA) in the subsequent \Cref{sec:kea}.

For our purposes, we consider strictly positive definite continuous kernels $k: \Omega \subset \Omega \rightarrow \R$, 
which are defined on some bounded subset $\Omega \subset \R^d$.
Strictly positive definite means, that the kernel matrix $k(X, X) := \left( k(x_i, x_j) \right)_{i,j=1}^n \subset \R^{n \times n}$ is positive definite for any choice of pairwise distinct points $\{x_1, ..., x_n \}_{i=1}^n \subset \Omega$.
Given such a kernel $k$, 
there always exists a unique native space of functions associated to $k$, the so called reproducing kernel Hilbert space $\ns \subset \mathcal{C}(\Omega)$. 
A typical example of such a kernel is given by the basic Matérn kernel
\begin{align}
\label{eq:basic_matern}
k(x, z) = \exp(-\Vert x - z \Vert),
\end{align}
which is also called exponential kernel or Laplace kernel.
The class of Matérn kernels will be used for the numerical experiments in \Cref{sec:numerical_experiments}.

Given a function $f \in \ns$ and pairwise distinct interpolation nodes $X_n \subset \Omega$, 
the kernel representer theorem states that there exists a minimum norm interpolant
\begin{align}
\label{eq:interpolant}
s_{f, X_n} = \sum_{j=1}^n \alpha_j^{(n)} k(\cdot, x_j),
\end{align}
where the coefficients $\alpha = ( \alpha_j^{(n)} )_{j=1}^n \subset \R^n$ can be computed directly by solving the linear equation system $k(X, X)\alpha = (f(x_i))_{i=1}^n$.
Thus accuracy of the interpolant $s_{f, X_n}$ for approximation of the target function $f$ crucially depends on the choice of the kernel and on the choice of interpolation points $\{ x_j \}_{j=1}^n \subset \Omega$.
In this work we deal with the second case, i.e.\ a suitable choice of interpolation points.
As elaborated in \Cref{sec:introduction}, 
greedy algorithms provide a computational efficient method to obtain a suitable set of interpolation points.

\subsection{Greedy point insertion}
\label{subsec:greedy_insertion}

Greedy insertion algorithms usually start with an empty set $X_0 := \{ \}$, 
which is iteratively updated by inserting a bunch of points.
We focus on the most popular case, which adds single points $x_{n+1}$ and thus reads $X_{n+1} := X_n \cup \{ x_{n+1} \}$.
This procedure is iterated, until a suitable expansion size $n$ is met or some accuracy or stability threshold is reached.
For the choice of the new point $x_{n+1}$, 
several criteria have been established (see e.g.\ \cite[Section 1]{wenzel2023analysis} for a more detailed discussion), 
of which we focus in the following on the residual based $f$-greedy and the power function based $P$-greedy criterion:
\begin{align}
\label{eq:selection_criteria}
\begin{aligned}
x_{n+1} &= \argmax_{x \in \Omega} |(f - s_{f, X_n})(x)| \qquad \qquad &&(f-\text{greedy}), \\
x_{n+1} &= \argmax_{x \in \Omega} P_{X_n}(x) && (P-\text{greedy}).
\end{aligned}
\end{align}
Here, $P_{X_n}(x)$ is the so-called power function, defined as 
\begin{align*}
P_{X_n}(x) = \sup_{0 \neq f \in \ns} \frac{|(f-s_{f, X_n})(x)|}{\Vert f \Vert_{\ns}},
\end{align*}
which measures the worst case error.
The power function can be computed efficiently based on the centers $X_n$ and the kernel $k$.
In practice, a large discrete base set $X \subset \Omega$ is used instead of the domain $\Omega$.

In order to update the kernel model Eq.~\eqref{eq:interpolant}, one typically does not use the kernel basis $\{ k(\cdot, x_j), j=1, ..., n \}$,
because it would require a recomputation of the coefficients $\{ \alpha_j^{(n)} \}_{j=1}^n$ for updating $s_{f, X_n}$ to $s_{f, X_{n+1}}$.
Therefore one usually prefers to work in the Newton basis,
which allows for efficient updating of $s_{f, X_n}$ to $s_{f, X_{n+1}}$.
We refer to the reference \cite{pazouki2011bases} for more updates on the Newton basis and the corresponding properties.
These efficient update procedures due to the Newton basis will also be leveraged for an efficient implementation of the kernel exchange algorithm (KEA), to be introduced in \Cref{alg:KEA}.

An implementation of such greedy insertion algorithms is provided e.g.\ by the VKOGA (\textit{vectorial kernel orthogonal greedy algorithm}) package \cite{santin2021kernel}.

\subsection{Greedy point removal}
\label{subsec:greedy_removal}

Greedy removal algorithms start with a large discrete base set $X \subset \Omega$,
and are iteratively updated by removing a subset of the included points \cite{marchetti2022efficient}.
In the following we consider the case of removing single points,
such that the update is given as $X_n := X_{n+1} \setminus \{ x_{n+1} \}$ for $n = |X|-1, |X| - 2, ...$\ .
This is iterated until a desired expansion size is reached or some accuracy threshold is met.

Analogously to the selection criteria of the greedy insertion algorithms in Eq.~\eqref{eq:selection_criteria},
there is again a residual based as well as a power function based criterion.
The idea is to remove the point which results in the smallest increase of the corresponding error indicator.
Thus in this case, the residual based as well as power function based selection criteria read
\begin{align}
\label{eq:removal_criteria}
\begin{aligned}
x_{n+1} &= \argmin_{x \in X_{n+1}} |(f - s_{f, X_{n+1} \setminus \{ x \}})(x)| \\
x_{n+1} &= \argmin_{x \in X_{n+1}} P_{{X_{n+1} \setminus \{ x \}}}(x).
\end{aligned}
\end{align}
Both these selection criteria are based on leave-one-out cross validation errors.
While the computation of these leave-one-out cross validation errors is computational more demanding,
there are efficient implementations based on \textit{Rippa's rule} and extensions thereof \cite{marchetti2021extension, rippa1999algorithm}.

An implementation of such greedy removal algorithms is provided e.g.\ by the ERBA (\textit{efficient reduced basis algorithm}) package \cite{marchetti2022efficient}. \\

Both greedy insertion algorithms as well as greedy removal algorithms are limited in the sense,
that they only increase respectively decrease the number of centers.
Thus, a suboptimal step can never be reversed, which can be seen as a limitation.
This limitation is lifted with the kernel exchange algorithms (KEA) introduced in \Cref{sec:kea},
as they exchange selected centers by inserting a center and also removing a center in every step,
thus performing a locally optimal update.

\section{Kernel exchange algorithm: KEA} 
\label{sec:kea}

\Cref{sec:introduction} and \Cref{sec:review_greedy} discussed several aspects of the optimality of the greedy insertion and removal algorithms.
Especially the convergence rates (in the number of interpolation points) of greedy insertion algorithms is known to be asymptocally optimal in several cases.
Nevertheless, the globally \textit{optimal} selection of interpolation points still remains unclear, especially due to its computational complexity. 
In order to narrow this gap between greedily selected points and optimal points from a practical point of view, 
we propose kernel exchange algorithms:

We consider an initial base set $(X, Y)$ of $N := |X|$ input points with corresponding target values $Y$,
as well as a kernel $k$ for approximation of these data points.
We assume a non-empty initial set $X_n \subset X$ of $n < N$ centers to be given, 
which can be obtained for example by a greedy insertion algorithm (see \Cref{subsec:greedy_insertion}) or a greedy removal algorithm (see \Cref{subsec:greedy_removal}).
Given a maximal number of $m$ exchange steps, for every exchange step $i = 1, ..., m$, 
we pick a data point of $X \setminus X_n$ to be added, 
as well as a data point of $X_n$ to be removed.
Like this, the set of selected centers is updated as
\begin{align}
\label{eq:update_step}
X_n^{(i+1)} := X^{(i)}_n \setminus \{ x_\text{remove} \} \cup \{ x_\text{add} \}.
\end{align}
For the selection of the points $x_\text{add}$ and $x_\text{remove}$, 
we leverage the residual based $f$-greedy criterion as well as the power function based $P$-greedy criterion, 
see Eq.~\eqref{eq:selection_criteria} respective Eq.~\eqref{eq:removal_criteria}.
The step of Eq.~\eqref{eq:update_step} is repeated, until the predefined number of maximal exchanges $m$ is reached, 
or some predefined stopping criterion (based e.g.\ on the final accuracy) is met.
The overall algorithm is formalized as pseudocode in \Cref{alg:KEA}.

For the implementation of the kernel exchange algorithm, 
we combine the efficient implementations of the greedy insertion and removal algorithms, 
as implemented e.g.\ in the algorithms VKOGA \cite{santin2021kernel} and ERBA \cite{marchetti2022efficient}, 
see \Cref{sec:review_greedy}.

\SetKwComment{Comment}{/* }{ */}

\RestyleAlgo{ruled}
\begin{algorithm}[hbt!]
\caption{Kernel Exchange Algorithm (KEA) for kernel model $\sum_{j=1}^n \alpha_j k(\cdot, x_j)$ of expansion size $n$}
\label{alg:KEA}
\SetKwInOut{Input}{Input}
\SetKwInOut{Output}{Output}
\Input{Data $(X, Y)$, initial set of centers $X^{(0)}_n \subset X$ of $n > 0$ points, \\ 
kernel $k$, number of exchange steps $m \in \N$}
\KwResult{Exchanged set of centers $X^{(m)}_n \subset X$, final kernel model $s^{(m)}_n(\tilde{X}_n)$} ~ \\

$s_0 = s_0(X_n^{(0)})$ \Comment*[r]{Compute initial kernel model}

~ \\
\For{$i = 1, ..., m$}{
	$x_\text{add} ~~~ \gets \textsc{select}\_\textsc{add}(s_n, X, Y)$ according to Eq.~\eqref{eq:selection_criteria} \;
	$x_\text{remove} \gets \textsc{select}\_\textsc{remove}(s_n, X, Y)$ according to Eq.~\eqref{eq:removal_criteria}\;
	$ X_n^{(i)} \gets X_n^{(i-1)} \setminus \{ x_\text{remove} \} \cup \{ x_\text{add} \}$\;
	~ \\
	
	$s_i = s_i(X_n^{(i)})$ \Comment*[r]{Compute updated kernel model} \
	
	\textsc{early}\_\textsc{stopping}($x_\text{add}$, $x_\text{remove}$)
}

~ \\
\textsc{return} $X_n^{(m)}$, $s_n = s_n(X_n^{(m)})$
\end{algorithm}

\section{Numerical experiments} \label{sec:numerical_experiments}

This section provides numerical experiments on the introduced kernel exchange algorithm (KEA) of \Cref{sec:kea}.
We start in \Cref{subsec:comparison} with a comparison of greedy insertion and greedy removal algorithms.
Subsequently, 
focussing on the case of greedy insertion algorithms, 
\Cref{subsec:low_dim_example} considers low dimensional and \Cref{subsec:high_dim_example} considers higher dimensional examples,
where KEA is used to finetune greedy kernel insertion algorithms.
In particular, \Cref{subsec:low_dim_example} and \Cref{subsec:high_dim_example} compare the accuracy of kernel models using a base set $X^{(0)}_n$ with the accuracy of a kernel model based on the exchanged set $X^{(m)}_n$ after using the KEA algorithm.

As kernels we consider Matérn kernels of different smoothnesses, namely $k(x, z) = \Phi(x-z)$,
where the function $\Phi: \R^d \rightarrow \R$ is defined via its Fourier transform 
$\hat{\Phi}(\omega) = (1 + \Vert \omega \Vert_2^2)^{-\tau}$
with $\tau = \frac{d+(2p+1)}{2}$ for $p \in \{0, 1, 2, 3, 4\}$.
All these kernels have an explizit form, and for $p=0$ we obtain the kernel from Eq.~\eqref{eq:basic_matern}.

The implementation of the KEA algorithm as well as the code to reproduce the numerical experiments can be found at: \\
\begin{center}
\url{https://gitlab.rrz.uni-hamburg.de/bbd9097/paper-2024-finetuning-greedy-kernel-models}
\end{center}
\vspace{.5cm}

\subsection{Greedy insertion vs.\ greedy removal}
\label{subsec:comparison}

In a first numerical experiment, we briefly compare the two possible approaches of greedy insertion of centers vs the greedy removal of centers as introduced in \Cref{subsec:greedy_insertion} and \Cref{subsec:greedy_removal}.
For this, we make use of the corresponding software packages VKOGA \cite{santin2021kernel} and ERBA \cite{marchetti2022efficient}.

\begin{figure}[h]
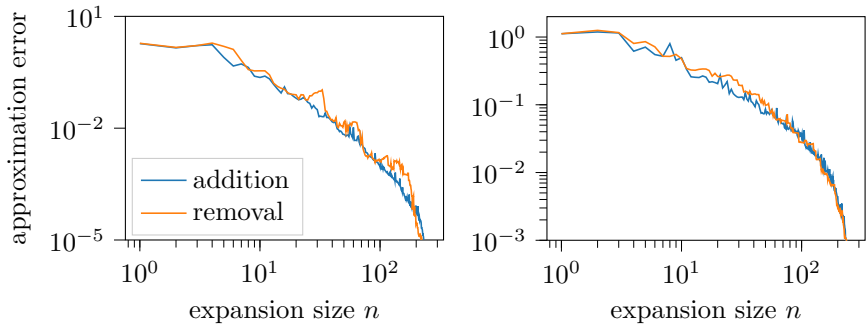

\centering
\setlength\fwidth{.5\textwidth}
\input{Figures/vkoga_vs_kea_visualization_0.tex}
\input{Figures/vkoga_vs_kea_visualization_1.tex}
\caption{
Visualization of the training error ($y$-axis) over the number of used interpolation points ($x$-axis) for the target functions in Eq.~\eqref{eq:two_target_functions}.
The insertion algorithm of \Cref{subsec:greedy_insertion} increases the number of points (operates from ``left to right'' on the $x$-axis),
the removal algorithm of \ref{subsec:greedy_removal} decreases the number of points (operates from ``right to left'').
Both approaches yield approximately the same interpolation errors for all expansion sizes $n$.
}
\label{fig:vkoga_vs_kea}
\end{figure}

We present two exemplary numerical results, though we remark that the findings presented here also hold for other examples.
We consider the domains $\Omega = [0, 1]^2$ respective $[0, 1]^3$ with target functions 
\begin{align}
\label{eq:two_target_functions}
f_2(x) = \Vert x \Vert^2 \qquad \text{respective} \qquad f_3(x) = |x_1 - 0.5| + \sin(x_2 + x_3).
\end{align}
As a base set, we consider each 256 low discrepancy points $X$ within $\Omega$.
In contrast to the numerical experiments in the next sections, we use low-discrepancy points instead of randomly sampled points to avoid numerical instabilities due to too small values,
which may occur as soon as nearby points are used as centers.
The greedy insertion algorithm starts with an empty set of centers, and adds centers until all the 256 points are used.
The removal algorithm operates in the reverse sense and starts with all the centers, and then removes centers until no center is used.
The resulting maximal error $\max_{x \in X} |f(x) - s_n(x)|$ over the number $n$ of centers is visualized for each the greedy insertion model and the greedy removal models $s_n$ in \Cref{fig:vkoga_vs_kea}:
One can observe that the maximal error is approximately equal for all the expansion sizes $n$.
Thus, from the theoretical point of view, either applying a greedy insertion or a greedy removal yields approximately the same accuracy.
However, from a practical point of view, if only a small number of centers $n \ll |X|$ is desired,
it makes more sense to leverage the greedy insertion algorithm.
On the contrary, if only few centers should be removed, i.e.\ $n \lessapprox |X|$, it is more practical to use the greedy removal algorithm.

For our motivated purpose of surrogate modeling, the first case $n \ll |X|$ is more important.
Thus we make use of the greedy insertion algorithm in the following,
and investigate to which extent its results can be improved by applying KEA (see \Cref{alg:KEA}).

\subsection{Function approximation: Low dimensional example}
\label{subsec:low_dim_example}

As a first test case, we consider the domain $\Omega = [0, 1]^2 \subset \R^d$ as input space and target values generated by four test functions.
For this we chose the classical well-known Franke test function $f_1$ as well as three further Franke test function given as \cite{franke1979critical}
\begin{align*}
f_2(x) ~ =& ~ \frac{1}{9} (\mathrm{tanh}(9x_2 - 9 x_1) + 1), \\
f_3(x) ~ =& ~ \frac{\frac{125}{100} + \cos(5.4 x_2)}{6 + 6(3 x_1 - 1)^2}, \\
f_4(x) ~ =& ~ \frac{1}{3} \exp \left( -\frac{81}{16} \left( (x_1 - 1/2)^2 + (x_2 - 1/2)^2 \right) \right).
\end{align*}

For all the four functions, the input domain $\Omega$ is discretized with each $10^3$ uniformly randomly sampled training points $X_\text{train}$ and testing points $X_\text{test}$.
The greedy insertion algorithm with the $f$-greedy criterion (see \Cref{subsec:greedy_insertion}) is used to select up to 150 (for $f_1$) respective 80 (for $f_2, f_3, f_4$) centers from the base set $X_\text{train}$.
For 10 logarithmically equally spaced values $n$ within 5 and 150 respective 80,
the resulting greedy insertion kernel model $s_n$ is finetuned with help of KEA (using at most $m=100$ exchange steps) to obtain the model $s_{n, \text{KEA}}$.

In order to assess the improvement due to the finetuning by KEA, we consider the improvement ratio on the test set $X_\text{test}$, i.e.\
\begin{align}
\label{eq:improvement_ratio}
\frac{\Vert f - s_{n, \text{KEA}} \Vert_{L^\infty(\Omega)}}{\Vert f - s_{n} \Vert_{L^\infty(\Omega)}},
\end{align}
computed on the $10^3$ test points.

\begin{figure}[h]
\centering
\setlength\fwidth{.5\textwidth}
\begin{tikzpicture}

\definecolor{blue015255}{RGB}{0,15,255}
\definecolor{cyan0255245}{RGB}{0,255,245}
\definecolor{darkgray176}{RGB}{176,176,176}
\definecolor{lightgray204}{RGB}{204,204,204}
\definecolor{lime72550}{RGB}{7,255,0}
\definecolor{yellow2512450}{RGB}{251,245,0}

\begin{axis}[
width=0.951\fwidth,
height=0.75\fwidth,
at={(0\fwidth,0\fwidth)},
legend cell align={left},
legend style={fill opacity=0.8, draw opacity=1, text opacity=1, draw=lightgray204},
log basis x={10},
tick align=outside,
tick pos=left,
x grid style={darkgray176},
xmin=4.21807154334655, xmax=177.806372483896,
xmode=log,
xtick style={color=black},
y grid style={darkgray176},
ymin=0, ymax=2,
ytick style={color=black}
]
\addplot [semithick, red, mark=x, mark size=3, mark options={solid}]
table {%
5 1
7 0.977399446081143
10 0.75723539148422
15 0.596380822638396
22 0.746660725151928
33 1.03041165464439
48 0.955242876440966
70 1.00873699499157
102 0.999999999999412
150 0.99997477031378
};
\addlegendentry{$p=0$}
\addplot [semithick, yellow2512450, mark=x, mark size=3, mark options={solid}]
table {%
5 1
7 1
10 1.00000000000002
15 0.754118214567615
22 0.766230737327174
33 0.795953574530136
48 0.725281911354508
70 0.765550772279908
102 0.90077030824664
150 1.00131193080744
};
\addlegendentry{$p=1$}
\addplot [semithick, lime72550, mark=x, mark size=3, mark options={solid}]
table {%
5 0.807237687317673
7 0.999999999999978
10 1.01581970230375
15 0.999999999999635
22 0.903463378987609
33 0.978633446419489
48 0.761578556468786
70 0.797778186717532
102 0.772783209711718
150 0.533973496251978
};
\addlegendentry{$p=2$}
\addplot [semithick, cyan0255245, mark=x, mark size=3, mark options={solid}]
table {%
5 0.688691615962704
7 1
10 0.714517909101754
15 0.999999999999539
22 0.801738807203608
33 0.999999999891599
48 0.507046542087897
70 0.565868065118874
102 0.857034509966711
150 0.582898398848128
};
\addlegendentry{$p=3$}
\addplot [semithick, blue015255, mark=x, mark size=3, mark options={solid}]
table {%
5 0.999999999999999
7 0.999999999999983
10 0.812887701081905
15 0.821188128466837
22 0.66189546832354
33 1.0000000000773
48 0.662236295475453
70 0.894352532912933
102 0.511259690925052
150 0.657245855970438
};
\addlegendentry{$p=4$}
\addplot [semithick, black, dashed, forget plot]
table {%
5 1
7 1
10 1
15 1
22 1
33 1
48 1
70 1
102 1
150 1
};
\end{axis}

\end{tikzpicture}
\begin{tikzpicture}

\definecolor{blue015255}{RGB}{0,15,255}
\definecolor{cyan0255245}{RGB}{0,255,245}
\definecolor{darkgray176}{RGB}{176,176,176}
\definecolor{lightgray204}{RGB}{204,204,204}
\definecolor{lime72550}{RGB}{7,255,0}
\definecolor{yellow2512450}{RGB}{251,245,0}

\begin{axis}[
width=0.951\fwidth,
height=0.75\fwidth,
at={(0\fwidth,0\fwidth)},
legend cell align={left},
legend style={fill opacity=0.8, draw opacity=1, text opacity=1, draw=lightgray204},
log basis x={10},
tick align=outside,
tick pos=left,
x grid style={darkgray176},
xmin=4, xmax=105,
xmode=log,
xtick style={color=black},
y grid style={darkgray176},
ymin=0, ymax=2,
ytick style={color=black}
]
\addplot [semithick, red, mark=x, mark size=3, mark options={solid}]
table {%
5 0.999999999999998
6 1
9 1
12 1.24257718683207
17 1.00000000000003
23 0.99977000772729
31 0.99999999999989
43 0.999999999999793
58 0.999783827685685
80 0.995220822745059
};
\addplot [semithick, yellow2512450, mark=x, mark size=3, mark options={solid}]
table {%
5 1
6 1
9 0.783028126400642
12 0.498659353092769
17 0.644343698263336
23 0.818868960667195
31 0.744858589387362
43 1.00491942649051
58 1.00108213443569
80 0.907930661966415
};
\addplot [semithick, lime72550, mark=x, mark size=3, mark options={solid}]
table {%
5 1.00000000000001
6 1
9 1.00000000000003
12 0.558122192489546
17 0.669769522744144
23 0.565696239402628
31 0.650560381149195
43 0.700663311873441
58 0.734114697125596
80 0.657779094857869
};
\addplot [semithick, cyan0255245, mark=x, mark size=3, mark options={solid}]
table {%
5 0.793477425715076
6 0.999999999999981
9 0.999999999999913
12 0.901348542692933
17 0.740827188532612
23 0.851705272224907
31 0.372852567803572
43 0.457793515656914
58 0.589345557077986
80 0.635945380226698
};
\addplot [semithick, blue015255, mark=x, mark size=3, mark options={solid}]
table {%
5 0.801488313837793
6 0.999999999999976
9 1.00000000000018
12 0.999999999999772
17 0.664162853240318
23 0.847351652347187
31 1.00000000174678
43 0.830623750830067
58 0.4676375946027
80 0.407423253560231
};
\addplot [semithick, black, dashed, forget plot]
table {%
5 1
6 1
9 1
12 1
17 1
23 1
31 1
43 1
58 1
80 1
};
\end{axis}

\end{tikzpicture}
\begin{tikzpicture}

\definecolor{blue015255}{RGB}{0,15,255}
\definecolor{cyan0255245}{RGB}{0,255,245}
\definecolor{darkgray176}{RGB}{176,176,176}
\definecolor{lightgray204}{RGB}{204,204,204}
\definecolor{lime72550}{RGB}{7,255,0}
\definecolor{yellow2512450}{RGB}{251,245,0}

\begin{axis}[
width=0.951\fwidth,
height=0.75\fwidth,
at={(0\fwidth,0\fwidth)},
legend cell align={left},
legend style={fill opacity=0.8, draw opacity=1, text opacity=1, draw=lightgray204},
log basis x={10},
tick align=outside,
tick pos=left,
x grid style={darkgray176},
xmin=4, xmax=105,
xmode=log,
xtick style={color=black},
y grid style={darkgray176},
ymin=0, ymax=2,
ytick style={color=black}
]
\addplot [semithick, red, mark=x, mark size=3, mark options={solid}]
table {%
5 1
6 1.00000000000001
9 1.00953147725597
12 1.00000000000002
17 1.00000000000005
23 1.00000000000008
31 0.999999999999836
43 0.999999999999693
58 0.999999999999118
80 1.0000000000017
};
\addplot [semithick, yellow2512450, mark=x, mark size=3, mark options={solid}]
table {%
5 1
6 0.999999999999996
9 0.919365359463672
12 0.722046367268417
17 1.00000000000105
23 0.835796750874173
31 0.631566119344207
43 0.735721991287933
58 0.990176128387829
80 1.01515944139907
};
\addplot [semithick, lime72550, mark=x, mark size=3, mark options={solid}]
table {%
5 0.999999999999991
6 0.685364557825931
9 1.0000000000001
12 0.136067527940849
17 1.00000000000389
23 1.31295693402422
31 0.91600227635157
43 0.593029101016484
58 0.473654312883433
80 0.622088298915312
};
\addplot [semithick, cyan0255245, mark=x, mark size=3, mark options={solid}]
table {%
5 0.999999999999996
6 1.00000000000001
9 0.99999999999982
12 0.193739601358222
17 0.555295514337961
23 0.71071238192909
31 0.999999979706333
43 0.477055240965898
58 0.99704762873141
80 0.988052731637444
};
\addplot [semithick, blue015255, mark=x, mark size=3, mark options={solid}]
table {%
5 0.999999999999992
6 1.00000000000001
9 0.99999999999995
12 0.67367752258185
17 1.04767677152245
23 0.625453893968987
31 0.564704515120627
43 0.772685173207682
58 0.973975462259139
80 0.64760199011817
};
\addplot [semithick, black, dashed, forget plot]
table {%
5 1
6 1
9 1
12 1
17 1
23 1
31 1
43 1
58 1
80 1
};
\end{axis}

\end{tikzpicture}
\begin{tikzpicture}

\definecolor{blue015255}{RGB}{0,15,255}
\definecolor{cyan0255245}{RGB}{0,255,245}
\definecolor{darkgray176}{RGB}{176,176,176}
\definecolor{lightgray204}{RGB}{204,204,204}
\definecolor{lime72550}{RGB}{7,255,0}
\definecolor{yellow2512450}{RGB}{251,245,0}

\begin{axis}[
width=0.951\fwidth,
height=0.75\fwidth,
at={(0\fwidth,0\fwidth)},
legend cell align={left},
legend style={fill opacity=0.8, draw opacity=1, text opacity=1, draw=lightgray204},
log basis x={10},
tick align=outside,
tick pos=left,
x grid style={darkgray176},
xmin=4, xmax=105,
xmode=log,
xtick style={color=black},
y grid style={darkgray176},
ymin=0, ymax=2,
ytick style={color=black}
]
\addplot [semithick, red, mark=x, mark size=3, mark options={solid}]
table {%
5 0.999999999999995
6 0.991885535161042
9 0.883379707772191
12 0.987329708788055
17 0.975182752196937
23 1.07105340497543
31 0.880127553310843
43 1.00000000000138
58 0.811306744533982
80 0.989359783696987
};
\addplot [semithick, yellow2512450, mark=x, mark size=3, mark options={solid}]
table {%
5 1.00000000000001
6 1.00000000000001
9 0.999999999999959
12 0.386152641977851
17 0.401545155162198
23 0.504350486390202
31 0.810298916706916
43 0.739321048878104
58 0.944251409587452
80 0.998894632847261
};
\addplot [semithick, lime72550, mark=x, mark size=3, mark options={solid}]
table {%
5 0.999999999999997
6 0.954626763990566
9 0.999999999999966
12 0.427257409978647
17 0.871529971630637
23 0.560769144702124
31 0.698537325403399
43 0.537685930481539
58 0.534451228269256
80 0.683879702450141
};
\addplot [semithick, cyan0255245, mark=x, mark size=3, mark options={solid}]
table {%
5 0.857926815085704
6 0.799011593652322
9 0.999999999999978
12 0.955542558513591
17 0.389819826264083
23 0.685737992435637
31 0.35965002271871
43 0.455273875259682
58 0.958537421719184
80 1.046427119886
};
\addplot [semithick, blue015255, mark=x, mark size=3, mark options={solid}]
table {%
5 1.00000000000001
6 0.755365450297324
9 0.999999999999925
12 0.644294100749974
17 0.959969287008587
23 0.289296803847983
31 0.682638716623524
43 0.509688551928473
58 0.328158973811272
80 0.797747583266755
};
\addplot [semithick, black, dashed, forget plot]
table {%
5 1
6 1
9 1
12 1
17 1
23 1
31 1
43 1
58 1
80 1
};
\end{axis}

\end{tikzpicture}
\caption{
Visualization of the improvement ratio $\Vert f - s_{n, \text{KEA}} \Vert_{L^\infty(\Omega)} / \Vert f - s_{n} \Vert_{L^\infty(\Omega)}$ ($y$-axis) over the kernel model expansion size $n$ ($x$-axis) for the four two-dimensional test functions from \Cref{subsec:low_dim_example}:
For values in $(0, 1)$, KEA yields improvements; for values in $(1, \infty)$, KEA yields deterioration.
Five Matérn kernels with different smoothness parameters $p \in \{0, 1, 2, 3, 4 \}$ were used.
}
\label{fig:vis_low_dim}
\end{figure}
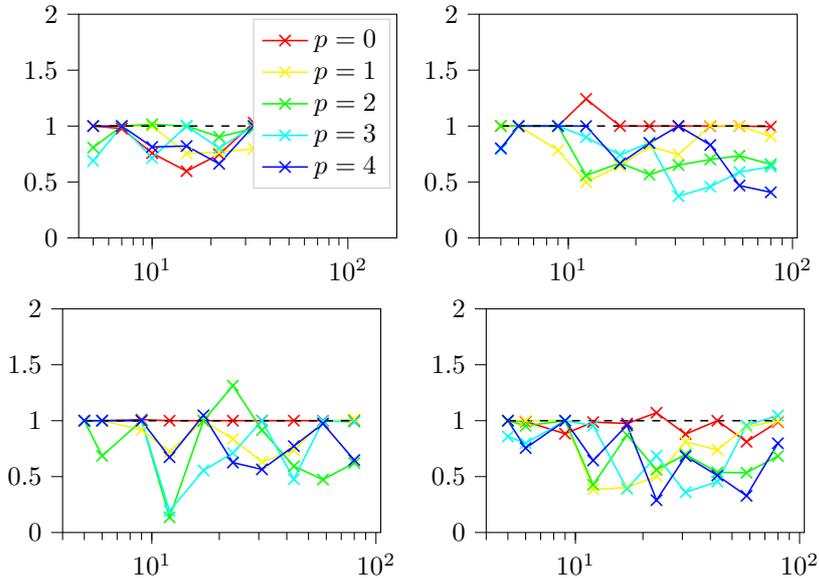

The resulting improvement ratios of Eq.~\eqref{eq:improvement_ratio} over the number of centers $n$ are visualized in \Cref{fig:vis_low_dim} for all the four test functions $f_1$ to $f_4$ and all the five considered Matérn kernels $k$.
It can be clearly seen, 
that in most cases the improvement ratio is below 1, which means that the finetuning of the kernel model $s_n$ due to KEA achieved a smaller test error.
The best improvement is obtained for the Matérn kernel with smoothness $p=2$ and an expansion size of $n=12$: 
Here, the improvement ratio is  $0.136$, which means that the exchange due to KEA gave a tremendous improvement,
without changing the size $n$ of the kernel model.
On average, an improvement of $17.2\%$ is obtained.
In general, the improvement seems to be more pronounced for smoother kernels, i.e.\ higher values of $p$.
This observation can be explained in view of the convergence analysis for greedy insertion algorithms in \cite{wenzel2023analysis}: 
There, a convergence bound as $C \cdot n^{-\alpha}$ (for some prefactor $C > 0$ and some convergence rate $\alpha > 0$) is proven.
The prefactor $C$ is increased (compared to the prefactor of a non-greedy algorithm) by an $\alpha$-dependent factors as
\begin{align}
\label{eq:constant_C}
C \propto 2^{\alpha + 1/2} e^\alpha,
\end{align}
see \cite[Corollary 11]{wenzel2023analysis}.
In the case of Matérn kernels considered here, it holds $\alpha = \frac{2p+1}{2d}$,
such that larger values of $p$ imply a larger increase of the prefactor $C$.
In effect, for larger values of $p$ there is a bigger gap, which can be narrowed with help of KEA.

Only for a few instances within \Cref{fig:vis_low_dim}, 
no improvement or in same rare cases even a deterioration can be observed.
We remark that such a deterioration can be observed despite the local optimality of the kernel exchange algorithm,
because the error is evaluated on an independent test set, which was withheld from the training process.

\subsection{Function approximation: High dimensional example}
\label{subsec:high_dim_example}

As a second test case, we consider the domains $\Omega = [0, 1]^d \subset \R^d$ for $d \in \{5, 6\}$ and the following two test functions, 
which were also used in \cite{wenzel2024data} as test functions for greedy approximation:
\begin{align*}
f_5(x) &= e^{-4 \left( \sum_{j=1}^5 \boldsymbol{x}_i - 0.5 \right)^2} \\
f_6(x) &= e^{-4 \sum_{j=1}^5 (\boldsymbol{x}_i - 0.5)^2} + 2 |x_1 - 0.5| %
\end{align*}

Again we test the five Matérn kernels from \Cref{subsec:low_dim_example}, 
however additionally making use of a two-layered kernel structure as $k(Ax, Az) = \Phi(A(x-z))$,
with a matrix $A \in \R^{d \times d}$ that is optimized.
This matrix $A$ allows to adapt the shape of the initially radial kernel $k$ to the data to be approximated, 
which usually improves the accuracy for medium- to high-dimensional problems.
For details on two-layered kernels and the corresponding optimization procedure to obtain a suitable matrix $A$, 
we refer to \cite{wenzel2024data}.

\begin{figure}[h]
\centering
\setlength\fwidth{.5\textwidth}
\begin{tikzpicture}

\definecolor{blue015255}{RGB}{0,15,255}
\definecolor{cyan0255245}{RGB}{0,255,245}
\definecolor{darkgray176}{RGB}{176,176,176}
\definecolor{lightgray204}{RGB}{204,204,204}
\definecolor{lime72550}{RGB}{7,255,0}
\definecolor{yellow2512450}{RGB}{251,245,0}

\begin{axis}[
width=0.951\fwidth,
height=0.75\fwidth,
at={(0\fwidth,0\fwidth)},
legend cell align={left},
legend style={fill opacity=0.8, draw opacity=1, text opacity=1, draw=lightgray204},
log basis x={10},
tick align=outside,
tick pos=left,
x grid style={darkgray176},
xmin=4, xmax=120,
xmode=log,
xtick style={color=black},
y grid style={darkgray176},
ymin=0, ymax=2,
ytick style={color=black}
]
\addplot [semithick, red, mark=x, mark size=3, mark options={solid}]
table {%
5 0.999999999999999
6 1
9 0.999999999999997
13 1.00000000000009
18 1.00000000000024
26 1.00117076210869
36 0.993798076343568
51 1.00000000001281
71 1.00000000004491
100 1.02797391768782
};
\addplot [semithick, yellow2512450, mark=x, mark size=3, mark options={solid}]
table {%
5 1.00000000000001
6 1
9 0.320751950133664
13 0.609937835245253
18 0.597956497352683
26 0.814381427979559
36 0.834924701617626
51 1.04241237043304
71 0.937901482377835
100 0.40208306722127
};
\addplot [semithick, lime72550, mark=x, mark size=3, mark options={solid}]
table {%
5 0.999999999999995
6 1
9 0.948173506613903
13 0.875960728895362
18 0.750486444375172
26 0.706655371240415
36 0.778784511635925
51 0.663187900699473
71 0.615781416317409
100 0.517565804496433
};
\addplot [semithick, cyan0255245, mark=x, mark size=3, mark options={solid}]
table {%
5 0.999999999999999
6 1
9 1.00000000000025
13 0.460446466158743
18 0.536978830122869
26 0.202117572505393
36 0.484505257459727
51 0.675210070074987
71 0.453947389993693
100 0.549088317435959
};
\addplot [semithick, blue015255, mark=x, mark size=3, mark options={solid}]
table {%
5 1
6 0.999999999999999
9 0.239086455675893
13 0.511403276212715
18 0.477148097520988
26 0.461699314435006
36 0.452579980099627
51 1.28341021528943
71 1.00254087385905
100 0.683589285965228
};
\addplot [semithick, black, dashed, forget plot]
table {%
5 1
6 1
9 1
13 1
18 1
26 1
36 1
51 1
71 1
100 1
};
\end{axis}

\end{tikzpicture}
\begin{tikzpicture}

\definecolor{blue015255}{RGB}{0,15,255}
\definecolor{cyan0255245}{RGB}{0,255,245}
\definecolor{darkgray176}{RGB}{176,176,176}
\definecolor{lightgray204}{RGB}{204,204,204}
\definecolor{lime72550}{RGB}{7,255,0}
\definecolor{yellow2512450}{RGB}{251,245,0}

\begin{axis}[
width=0.951\fwidth,
height=0.75\fwidth,
at={(0\fwidth,0\fwidth)},
legend cell align={left},
legend style={fill opacity=0.8, draw opacity=1, text opacity=1, draw=lightgray204},
log basis x={10},
tick align=outside,
tick pos=left,
x grid style={darkgray176},
xmin=4, xmax=250,
xmode=log,
xtick style={color=black},
y grid style={darkgray176},
ymin=0, ymax=2,
ytick style={color=black}
]
\addplot [semithick, red, mark=x, mark size=3, mark options={solid}]
table {%
5 0.999999999999999
7 1
11 1
17 0.999999999999998
25 0.951510046772157
38 0.911368419678112
58 1.01794657204906
88 1.02721117726487
132 0.970617609298274
200 0.995493623547264
};
\addplot [semithick, yellow2512450, mark=x, mark size=3, mark options={solid}]
table {%
5 0.792082595982679
7 0.789258996552368
11 1.00000000000012
17 0.666765701039031
25 0.674801835046581
38 1.01442445779472
58 0.805979341142709
88 0.824199252636651
132 0.949189561317186
200 1.05301860919442
};
\addplot [semithick, lime72550, mark=x, mark size=3, mark options={solid}]
table {%
5 0.446509994478285
7 0.631819068082744
11 0.895138284029144
17 1.09621380924926
25 0.452371012289212
38 1.00000000000887
58 0.669864914097052
88 0.537768507500596
132 1.0924212928668
200 0.995280834356952
};
\addplot [semithick, cyan0255245, mark=x, mark size=3, mark options={solid}]
table {%
5 0.249072282131798
7 0.52471216311762
11 0.841140887447402
17 0.604470472596109
25 0.768821801520191
38 1.05355413081374
58 0.654340524325388
88 0.788184507780972
132 0.769497558099385
200 0.551686224596768
};
\addplot [semithick, blue015255, mark=x, mark size=3, mark options={solid}]
table {%
5 0.249798979367202
7 0.350120724467697
11 1.00880259873791
17 0.714142631749204
25 0.506018942651368
38 0.582611264412262
58 0.884121759214498
88 0.453997283939733
132 0.81322649061713
200 0.624836442252295
};
\addplot [semithick, black, dashed, forget plot]
table {%
5 1
7 1
11 1
17 1
25 1
38 1
58 1
88 1
132 1
200 1
};
\end{axis}

\end{tikzpicture}
\caption{
Visualization of the improvement ratio $\Vert f - s_{n, \text{KEA}} \Vert_{L^\infty(\Omega)} / \Vert f - s_{n} \Vert_{L^\infty(\Omega)}$ ($y$-axis) over the kernel model expansion size $n$ ($x$-axis) for the two high-dimensional test functions from \Cref{subsec:high_dim_example}:
For values in $(0, 1)$, KEA yields improvements; for values in $(1, \infty)$, KEA yields deterioration.
Five Matérn kernels with different smoothness parameters $p \in \{0, 1, 2, 3, 4 \}$ were used.
}
\label{fig:vis_high_dim}
\end{figure}
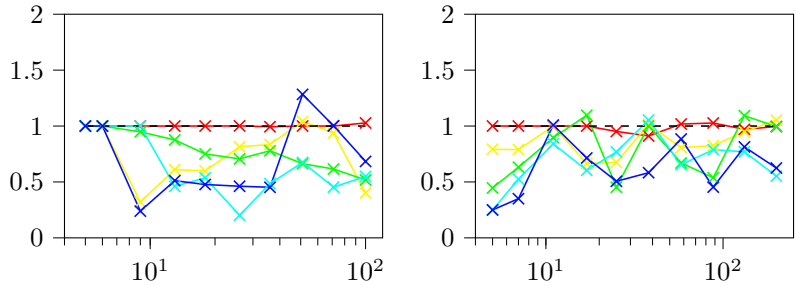

The layout of the numerical experiment is the same as previously in \Cref{subsec:low_dim_example}, 
with a couple of minor changes:
In order to take into account the higher dimensionality of the domain $\Omega$, 
we employ $10^4$ uniformly randomly sampled points for $X_\text{train}$ as well as $X_\text{test}$. 
The greedy insertion algorithm uses again the $f$-greedy criterion, 
and selects up to 100 centers for $f_5$ respective 200 centers for $f_6$.
KEA is applied again to intermediate models of size $n$ for 10 logarithmically equally spaced values of $n$ between $5$ and the maximal expansion size 100 respective 200.
The improvement ratio of Eq.~\eqref{eq:improvement_ratio} is considered, and the results are displayed in \Cref{fig:vis_high_dim}:

As in the low dimensional examples, 
one can observe that the use of KEA further reduces the approximation error for most expansion sizes as well as most kernels. 
The improvement is more pronounciated for smoother kernels (i.e.\ large values of $p$),
especially for $p=0$ there is frequently no improvement.
The same explanation as given around Eq.~\eqref{eq:constant_C} also applies here.
Only in rare cases, there is a deterioration instead of an improvement of the ratio Eq.~\eqref{eq:improvement_ratio}.

\section{Conclusion \& Outlook} \label{sec:conclusion_outlook}

In this work, two approaches of greedy kernel algorithms for interpolation were considered and compared, namely greedy insertion and greedy removal of points. 
The driving motivation for these algorithms is to derive sparse and efficient kernel models in a computational feasible way.

In order to finetune these greedy kernel models, we introduced and investigated a \textit{kernel exchange algorithm} (KEA):
Based on an initial set of centers, provided e.g.\ by a greedy algorithm,
exchange steps are performed to further optimize the selected subset of centers, 
without increasing (or decreasing) the amount of centers.
Doing so, 
we showed that it is indeed possible to further improve the accuracy of the final kernel model.

Future work may address the quantification of the possible improvement, in particular in comparison to a theoretically optimal center distribution, 
which is however computational infeasible in most cases. \\

\textbf{Acknowledgements:}
The authors acknowledge financial support through the projects LD-SODA of the {\em Landesforschungsf\"orderung Hamburg} (LFF)
and support from the RTG~2583 ``Modeling, Simulation and Optimization of Fluid Dynamic Applications''
funded by the {\em Deutsche Forschungsgemeinschaft} (DFG).

\bibliography{/home/wenzel/references}				%
\bibliographystyle{abbrv}

\end{document}